%% file: main.tex
\PassOptionsToPackage{table}{xcolor}
\documentclass[a4paper, twocolumn]{article}

\usepackage{preprint}

\usepackage[breaklinks=true]{hyperref}
\hypersetup{colorlinks,allcolors=black}
\usepackage{breakcites}

\usepackage{amssymb,amsmath,array,bm}
\usepackage{booktabs}
\usepackage{tabularx}
\newcolumntype{Y}{>{\centering\arraybackslash}X}

\usepackage{tikz}
\usetikzlibrary{arrows}

\usepackage{pgfplots}
\pgfplotsset{compat=1.15}
\usetikzlibrary{arrows}

\usepackage[a4paper, left=2.2cm, right=2.2cm, top=3cm, bottom=3cm]{geometry}

\usepackage{authblk}
\usepackage[numbers]{natbib}
\usepackage{doi}
\usepackage{orcidlink}

\usepackage[utf8]{inputenc}
\usepackage[T1]{fontenc}
\usepackage[english]{babel}
\usepackage{csquotes}

\usepackage{multirow}
\usepackage{makecell}
\usepackage{rotating}
\usepackage{colortbl}
\usepackage{subcaption}
\usepackage{float}
\usepackage{xcolor}
\definecolor{humanbg}{RGB}{235,242,250} 
\definecolor{llmbg}{RGB}{245,238,230} 
\definecolor{replace_old}{HTML}{FF7F50}
\definecolor{replace_new}{HTML}{008B8B}
\definecolor{insert}{HTML}{228B22}
\definecolor{delete}{HTML}{DC143C}
\usepackage{framed}
\usepackage{caption}
\usepackage{graphics}
\usepackage{enumitem}

\title{Limits of LLM Text Detectors in Education}

\author[1]{Lukas Gehring\orcidlink{0009-0009-3335-1679}}
\author[1]{Benjamin Paaßen\orcidlink{0000-0002-3899-2450}}
\affil[1]{Faculty of Technology, Bielefeld University}

\date{preprint as provided by the authors}

\begin{document}

\twocolumn[
    \begin{@twocolumnfalse} 
  
    \maketitle

    \input{sections/00-abstract}

    \vspace{0.35cm}

    \end{@twocolumnfalse} 
]
\footnotetext[1]{\href{https://github.com/lukasgehring/Assessing-LLM-Text-Detection-in-Educational-Contexts}{Online supplementary available at: https://github.com/lukasgehring/Assessing-LLM-Text-Detection-in-Educational-Contexts}}
\setcounter{footnote}{1}
\input{sections/01-introduction}

\input{sections/02-related_work}

\input{sections/03-method}

\input{sections/05-results}

\input{sections/06-discussion}

\section*{Acknowledgement}

We gratefully acknowledge funding for the project KI-Akademie OWL, financed by the Federal Ministry of Research, Technology and Space (BMFTR) and supported by the Project Management Agency of the German Aerospace Centre (DLR) under grant no. 01IS24057A.

\bibliographystyle{plainnat}
\bibliography{ref-short}

\appendix
\onecolumn
\input{sections/09-appendix}
\end{document}

%% file: sections/00-abstract.tex
\begin{abstract}
Students increasingly use the assistance of large language models (LLMs) in their academic writing. While slight assistance (e.g., grammar and style correction, as well as feedback) is permitted under most institutional policies, it is usually forbidden to offload entire writing tasks to LLMs. Unfortunately, current approaches to LLM-generated text detection predominantly assume a binary distinction between human-written and LLM-generated text, ignoring the breadth of realistic human-AI collaboration practices and limiting the validity of detection systems for educational assessment.
In this paper, we propose a contribution-aware evaluation framework for LLM-based detection systems in education. We introduce a scale of eight student contribution levels that model realistic writing scenarios ranging from fully human-written texts to LLM-assisted revisions to fully LLM-generated and adversarially humanized texts. Institutional policies regarding LLM use can then be translated to thresholds of acceptable LLM assistance on this scale.
We further present Generative Essay Detection in Education (GEDE), a novel benchmark dataset comprising more than 900 human-written and over 12,500 generated essays across 886 tasks and all contribution levels. 
Using this benchmark, we conduct a systematic evaluation of four state-of-the-art zero-shot and supervised detection methods across policy boundaries, contribution levels, generative models, out-of-distribution data, and text length.
We show that most detectors struggle to accurately classify texts at intermediate student contribution levels, in particular LLM-assisted revisions of human-written texts. Such errors pose a substantial risk of false accusations, indicating that current text detection systems are (still) unsuitable to reliably support the enforcement of institutional policies regarding LLM assistance in education.
Our dataset, code, and additional supplementary materials are publicly available\footnotemark[1].
\end{abstract}

%% file: sections/01-introduction.tex
\section{Introduction}

Large language models (LLMs) are rapidly transforming students' writing in education \cite{freeman_student_2025}. 
Students increasingly use LLMs not only to generate complete texts, but also to revise and structure drafts, improve grammar and style, or paraphrase texts.
In doing so, students employ a wide range of human-AI collaboration practices, yielding texts that are a mixture of human and LLM contributions. 
Such practices raise fundamental challenges for educational assessment and academic integrity~\cite{Kloke2024}. 
Writing tasks may no longer primarily assess students’ own skills, but rather their ability to use LLMs effectively, while authorship becomes increasingly difficult to determine (and define).
Consequently, institutions seek to ensure substantial student contribution and require ways to distinguish human-written from LLM-generated text.
However, manual detection remains unreliable even for experienced teachers~\cite{FLECKENSTEIN2024100209,liu2023argugpt}.
As a result, interest in automated methods for detecting LLM-generated text has grown substantially~\cite{Wu2025} and detection systems are increasingly integrated into educational settings~\cite{CDT2024AIDetectionEducation}.
To be reliable, detectors must operate robustly under realistic student writing behaviors and institutional usage policies. 
However, the practical validity of current detectors in such collaborative scenarios remains largely unexplored~\cite{Wu2025}.


Most existing work formulates LLM detection as a binary classification problem, distinguishing human-written from LLM-generated text. 
This assumption conflicts with institutional policies that typically permit some level of LLM use, while requiring substantial student contribution \cite{Wang2024}.
Therefore, it remains unclear how current detection methods behave across the spectrum of human-AI collaboration practices employed by students. 
We argue that LLM detection in educational settings must be evaluated in the context of \emph{student contribution}. 
Rather than treating texts as purely human-written or LLM-generated, we propose an ordinal scale of human and LLM contribution levels. Different institutional policies can, then, be translated into different thresholds of acceptable LLM usage on this scale.

\begin{figure*}[tb]
\centering
  \includegraphics[width=.95\textwidth]{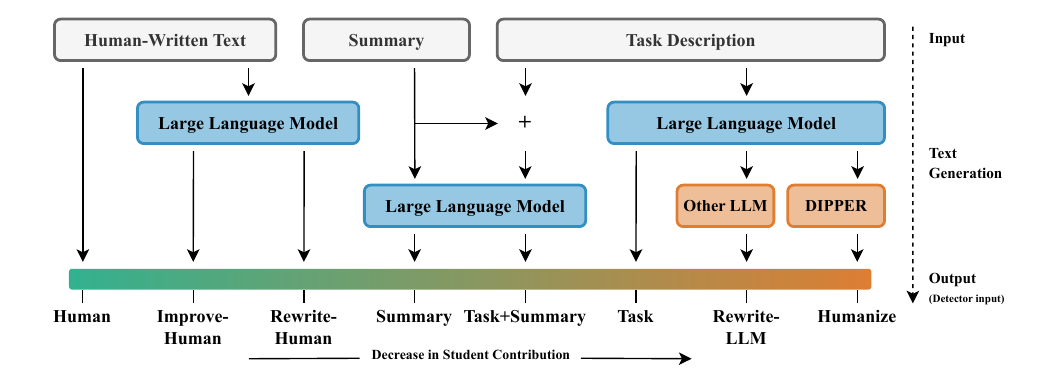}
  \caption{Overview of the different contribution levels, from fully human-written to fully generated texts, and how the different contribution levels are generated.}
  \label{fig:method-overview}
\end{figure*}
This paper makes four main contributions.
First, we introduce a scale of eight \emph{contribution levels} that capture realistic writing practices, ranging from entirely human-written text to LLM-assisted revision and summary-based generation to fully LLM-generated text based on task descriptions and texts where the use of LLMs is actively concealed or ``humanized'' (see Figure~\ref{fig:method-overview}).
Second, we provide the first systematic assessment of LLM detection performance across this broad spectrum of contribution levels and formalize institutional usage policies as \emph{label boundaries} that determine which contribution levels are considered acceptable.
Third, we introduce the Generative Essay Detection in Education (GEDE) benchmark dataset, comprising more than 900 human-written and over 12,500 LLM-generated essays across 886 unique task descriptions.
Finally, we conduct a comprehensive evaluation of four state-of-the-art zero-shot and supervised detection methods to assess the impact of policy boundaries, contribution levels, generative models, out-of-distribution data, and text length on the detectors’ performance.

Our results show that, while some detectors achieve high scores for fully LLM-generated texts, performance collapses at intermediate contribution levels, particularly LLM-assisted revisions of human-written texts. These findings indicate that current detection systems are unsuitable to implement the LLM usage policies of educational institutions -- because they do not accurately measure human contribution.

%% file: sections/02-related_work.tex
\section{Related Work}
\label{sec:related-work}

\subsection{Automated Detection of LLM-Generated Text}
\label{sec:detection-methods}
Precursors to the detection of LLM-generated texts include stylometric analysis and authorship attribution, where lexical diversity, word frequencies, and other stylistic markers serve as features for distinguishing text provenance across sources~\cite{Stamatatos2009}. 
Early LLM-generated text detection methods, such as GLTR~\cite{gehrmann2019gltr}, operationalize this idea by analyzing token-level probability ranks and local regularities to distinguish human from LLM-generated text. 

These \textbf{zero-shot} methods do not require training on labeled texts and instead rely on predefined indicators derived from language model statistics, such as token likelihoods, entropy, or perplexity.
DetectGPT~\cite{Mitchell2023} is a prominent zero-shot detector that exploits the tendency of LLM outputs to lie near local maxima of the model's log probability function. It classifies text by comparing the model likelihood of the original text to that of small perturbations, labeling texts as LLM-generated when the original has a higher likelihood than its perturbed variants. 
Fast-DetectGPT~\cite{bao2023fast} is a more efficient variant of DetectGPT that substitutes the perturbation step with token-level conditional probability scoring. 

\textbf{Supervised} detectors collect training data of human-written and LLM-generated text to train a machine learning model that distinguishes between both. Existing approaches can be broadly grouped into feature-based detectors and pre-trained text-embedding-based detectors~\cite {Wu2025}.
Feature-based detectors rely on explicit indicators derived from linguistic features~\cite{Shah2023}, model-based statistics~\cite{wang-etal-2023-seqxgpt}, or combinations of both~\cite{Mindner2023,verma-etal-2024-ghostbuster}.
Ghostbuster~\cite{verma-etal-2024-ghostbuster} is a supervised feature-based detector that applies a structured search over combinations of token probabilities from a series of weak learners together with handcrafted features, and trains a linear classifier on them.
A commonly used method of pre-trained detectors is to fine-tune a Transformer-based LM (e.g., BERT~\cite{devlin-etal-2019-bert} or RoBERTa~\cite{zhuang-etal-2021-robustly}), showing remarkable performance in many domains~\cite{He2024,liu2023argugpt,wang-etal-2024-m4}.

\textbf{Watermarking} methods embed patterns directly into LLM-generated text for identification and typically require access to the generation process~\cite{LiuACMComput2024}. In educational settings, its practical relevance is currently limited, as watermarks can often be weakened by text editing and are not implemented in widely deployed LLMs. Therefore, we do not consider watermarking techniques in this study.

\subsection{Benchmark Datasets for LLM Text Detection}

Most research on LLM-generated text detection mainly frames the problem as a binary classification task, distinguishing human-written from LLM-generated text. 
General-domain benchmarks such as TuringBench~\cite{uchendu-etal-2021-turingbench-benchmark} and M4~\cite{wang-etal-2024-m4} label texts as either human-written or LLM-generated and evaluate detectors accordingly, revealing important limitations in cross-domain and cross-model generalization.
MGTBench~\cite{He2024} and ArguGPT~\cite{liu2023argugpt} contain argumentative student essays and, thus, extend this evaluation to student writing and demonstrate the difficulty of manual detection and the brittleness of existing detectors. 
Evaluating detectors on real student submissions, Orenstrakh et al.~\cite{orenstrakh2024} further highlight false-positive risks that constrain their institutional deployment. 
However, these works continue to assume a strict binary authorship and do not examine intermediate forms of human–AI collaboration.

Dou et al.~\cite{dou2024} show that detection accuracy is highly sensitive to the content of the input, suggesting that the robustness of the detector depends not only on the generated text but also on the content provided to the LLM. Still, a multi-level model of student contribution is missing.

A small number of studies explicitly address collaborative human–AI authorship in text detection. Zhang et al.~\cite{zhang2024llm-as-coauthor} (MIXSET) introduce mixed human–AI texts and evaluate detector robustness under various editing operations, demonstrating that existing detectors struggle with subtle revisions and humanization. He et al.~\cite{he2025detree} (concurrently to our work) construct RealBench, a large-scale benchmark for analyzing human–AI collaborative writing at scale, and propose a structured, hierarchical characterization of hybrid texts based on their generation processes.

While these works advance detection methodologies and the modeling of hybrid authorship, they conceptualize human–AI collaboration using discrete categories or structured relations between such categories, rather than a fine-grained ordinal scale of human contribution, that allow the definition of a boundary between acceptable and non-acceptable LLM usage. More importantly, these prior works are mostly focused on texts outside education and thus fail to take educational specificities into account.
In contrast, our work adopts an application-driven view and investigates the practical validity of existing detectors for educational assessment.
We analyze detector behavior across realistic levels of student contribution and institutional policy boundaries, with a focus on alignment with educational usage policies and implications for fair assessment.

%% file: sections/03-method.tex
\section{Method}
\label{sec:method}

To evaluate LLM detection under realistic educational writing practices, we propose the concept of \emph{student contribution}.  We operationalize this concept as an ordinal scale of eight \emph{contribution levels} and formalize institutional policies as \emph{label boundaries} (Sec.~\ref{sec:contribution-categoires}). Based on this framework, we construct the Generative Essay Detection in Education (GEDE) dataset for a systematic detector evaluation (Sec.~\ref{sec:dataset}).

\subsection{Student Contribution and Policy Boundaries}

\label{sec:contribution-categoires}

In educational writing contexts, students use LLMs in diverse ways that extend beyond fully generating texts, including enhancing or proofreading their work~\cite{freeman_student_2025,hirabayashi2024harvardundergraduatesurveygenerative}. To account for this diversity in student writing practices, we introduce the concept of \emph{student contribution}.
We operationalize student contribution by defining a scale of eight \emph{contribution levels} that represent common patterns of student-LLM interaction in educational writing:

\begin{itemize}[label=\small$\bullet$,topsep=0pt, itemsep=2pt] 
    \item \textbf{Human}: Essays written entirely by students without any LLM assistance.
    
    \item \textbf{Improve-Human}: Human-written essays with minor grammatical and stylistic corrections applied by an LLM, without altering content or structure.
    
    \item \textbf{Rewrite-Human}: Human-written essays rewritten by an LLM, allowing for broader stylistic and structural changes while preserving the original content.
    
    \item \textbf{Summary}: Essays generated by an LLM from bullet-point style summaries, approximating student-provided prompts. As such summaries are not available in our corpora, they are automatically derived from the corresponding human essays using T5~\cite{Raffel2020}.
    
    \item \textbf{Task and Summary}: Same as the previous level, but with the task description additionally provided in the prompt.
    
    \item \textbf{Task}: Essays generated solely from the task description, without any student-provided content.
    
    \item \textbf{Rewrite-LLM}: LLM-generated essays rewritten by a different LLM to simulate attempts to circumvent detectors.
    
    \item \textbf{Humanize}: Adversarial rewriting of LLM-generated essays using DIPPER~\cite{Krishna2023}, a discourse-aware paraphrasing model designed to conceal LLM usage.
\end{itemize}
Figure~\ref{fig:method-overview} provides an overview of all contribution levels and the corresponding generation procedures.

In educational practice, institutional policies determine which levels of contribution are considered acceptable and which constitute an unacceptable level of help by an LLM. We formalize this decision by choosing a boundary, separating acceptable from non-acceptable contribution levels. The former constitutes the negative class (acceptable) for binary classification; the latter constitutes the positive class (non-acceptable). We investigate detection performance across different contribution levels (Sec.~\ref{sec:detector_comparison}), and the impact of different boundary choices (Sec.~\ref{sec:label_boundary}).

\subsection{Generative Essay Detection in Education Dataset}
\label{sec:dataset}
This work introduces the Generative Essay Detection in Education (GEDE) dataset, which comprises fully human-written and LLM-generated essays across all eight contribution levels.

\subsubsection{Human-Data Collection:}
GEDE includes human-written essays predating the release of ChatGPT from three publicly available text corpora, serving as sources of realistic, student-written texts in educational contexts: the Argument Annotated Essay (AAE) dataset~\cite{Stab2017,stab-gurevych-2017-parsing}, the PERSUADE 2.0 corpus~\cite{Crossley2024}, and the BAWE corpus~\cite{BAWE2008}, each providing essays with corresponding task descriptions.
We collected a total of 916 human-written essays for 826 unique task descriptions from those corpora.

\begin{table*}[tb]
    \centering
    \scriptsize  
    \caption{Number of human-written and LLM-generated essays per corpus.}
    \label{tab:dataset-samples}
    \input{tables/dataset_overview}
\end{table*}

The Argument Annotated Essay (AAE) dataset~\cite{Stab2017,stab-gurevych-2017-parsing} is a corpus of argumentative essays published in 2017.  It contains 402 English student essays, collected from \textit{essayforum.com}, with an average length of approximately 300 words, many of which are based on questions from the IELTS\footnote{https://ielts.org/} language test, an internationally recognized English test.
The PERSUADE 2.0 corpus~\cite{Crossley2024} includes over 25,000 argumentative essays written by 6th-12th grade students in the United States, collected before the release of ChatGPT in November 2022. Unfortunately, these essays are based on only 15 task descriptions, limiting the diversity of the texts. Accordingly, we include only five (randomly selected) texts per task description, each approximately 300 words, resulting in 75 human-written texts.
The British Academic Written English (BAWE) corpus~\cite{BAWE2008} contains 2,761 texts across 35 disciplines, written by 1st–4th-year students at Oxford Brookes, Reading, and Warwick universities, collected before 2004. From each discipline, 25 texts were randomly selected (or all texts where fewer were available), resulting in 439 human-written texts.
To meet the detectors’ input limit, all texts were truncated at sentence boundaries to a maximum length of 320 words. Detailed statistics for all three corpora are provided in the appendix~\ref{sec:dataset_statistics}.

\subsubsection{LLM Text Generation:}
The GEDE dataset contains one LLM-generated essay per contribution level and task description from each of the three corpora introduced in the previous section. 
For each contribution level, we construct a level-specific prompt that includes all relevant inputs.
The online supplement provides all exact prompts and generation parameters, qualitative examples illustrating model-specific edits at the \textit{Improve-Human} contribution level, and a similarity analysis across the \textit{Human}, \textit{Improve-Human}, and \textit{Rewrite-Human} contribution levels\footnotemark[1].
Figure~\ref{fig:method-overview} shows the different inputs provided to the LLM during text generation.

Since OpenAI’s ChatGPT is the most widely used LLM among students~\cite{hirabayashi2024harvardundergraduatesurveygenerative}, we use GPT-4o-mini---the base model of the free ChatGPT version at the time of analysis---to generate LLM-written essays via the OpenAI API. To assess the impact of the generative model on detection performance, we additionally include Meta’s open-source Llama-3.3-70b, which is widely accessible in German academia. 
The Rewrite-LLM and Humanize levels constitute special cases, as the LLM output is rewritten by another LLM or by the DIPPER model, respectively.
For the PERSUADE corpus, we observed that prompting an LLM with the same task description multiple times yielded highly similar LLM-generated essays. Accordingly, following the approach of Zhuo et al. \cite{zhuo-etal-2024-prosa}, we generated four paraphrases of the task description using Llama-3.3-70b, yielding five unique essays per task. Overall, the dataset totals 5,460 essays for each generative model and 1,783 for the DIPPER model. The number of generated essays for each text corpus can be found in Table~\ref{tab:dataset-samples}. Minor variations in the number of generated texts are due to the maximum input size limitations of the T5 and DIPPER models.

%% file: tables/dataset_overview.tex
\begin{tabularx}{\textwidth}{l|Y|Y|Y|Y|Y|Y|Y|Y}
    
    \toprule
        Text Corpus & \makecell[bl]{Human}
& \makecell[bl]{Improve-\\Human}
& \makecell[bl]{Rewrite-\\Human}
& \makecell[bl]{Summary}
& \makecell[bl]{Task+\\Summary}
& \makecell[bl]{Task}
& \makecell[bl]{Rewrite-\\LLM}
& Humanize\\
        \midrule
        AAE &  402 & 804 & 804 & 758 & 758 & 804 & 804 & 804\\
        PERSUADE &  75 & 150 & 150 & 150 & 150 & 150 & 150 & 135 \\
        BAWE &  439 & 878 & 878 & 868 & 868 & 878 & 878 & 844 \\
        \bottomrule
    \end{tabularx}

%% file: sections/05-results.tex
\section{Experiments}
\label{sec:results}
We evaluate existing detection methods on the proposed GEDE dataset to assess performance across label boundaries (reflecting different LLM usage policies), contribution levels, generative models, out-of-distribution data, and text lengths.
We investigate four detection methods (Sec.~\ref{sec:detection-methods}): the zero-shot models DetectGPT and Fast-DetectGPT, the supervised model Ghostbuster, and a RoBERTa-based classifier. For Ghostbuster, we use the authors’ released pre-trained parameters. 
For RoBERTa, we fine-tune \textit{roberta-base} separately on each of the three corpora using only the \textit{Human} (negative class) and \textit{Task} (positive class) contribution levels, reflecting common detector training practices. To estimate performance on the entire dataset, cross-validation across corpora is employed (e.g., performance on AAE is evaluated using a RoBERTa model trained on BAWE or PERSUADE).
Additional training details are provided in the appendix Table~\ref{tab:roberta-training}.

We evaluate the detectors using two complementary metrics. ROC-AUC~\cite{FAWCETT2006861} measures the probability that a detector assigns higher scores to LLM-generated than to human-written texts~\cite{Fernández2018} and is widely used in prior work on LLM detection~\cite{bao2023fast,Mitchell2023,Tulchinskii2023}. While ROC-AUC is not fully robust to class imbalance, it generally provides reliable, threshold-independent performance estimates when a sufficient number of samples is available for the minority class~\cite{Fernández2018}.
We also report TPR@5\%FPR, following He et al.~\cite{he2025detree}, to assess detection performance under a strict false-positive constraint, reflecting the fact that educational institutions typically want to avoid falsely accusing students of cheating. ROC curves for all experiments are provided in the appendix~\ref{sec:roc_curves}

\subsection{Detector Performance Boundary Analysis}
\label{sec:label_boundary}
In practice, educators must decide which contribution levels are considered acceptable and which contribute an unacceptable amount of LLM support.
In our framework, this policy decision corresponds to the choice of a \emph{label boundary} that partitions the spectrum of contribution levels into a negative (acceptable) and a positive (non-acceptable) class.
We evaluate the sensitivity of detectors to different policy choices by systematically shifting the label boundary and measuring performance as increasingly LLM-involved contribution levels are considered acceptable. Specifically, the boundary indicates the last contribution level included in the negative class.

\begin{table*}[t]
    \centering
    \scriptsize
    \caption{ROC-AUC and TPR@5\%FPR scores for different label boundaries and detectors. The boundary indicates the last contribution level assigned to the negative (acceptable) class; all subsequent levels are treated as non-acceptable.}
    \label{tab:label_boundary_roc}
    \input{tables/human_label_boundary}
\end{table*}

\begin{figure*}[ht]
    \centering
    \includegraphics[width=.9\linewidth]{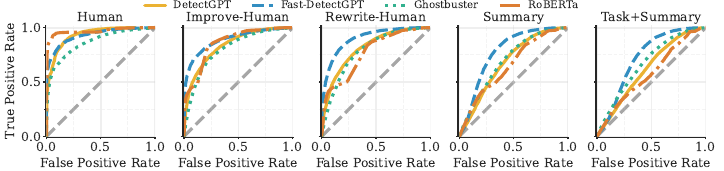}
    \caption{ROC curves at different label boundaries for all detectors.}
    \label{fig:boundary_roc}
\end{figure*}

Table~\ref{tab:label_boundary_roc} reports ROC-AUC and TPR@5\%FPR scores for five boundary choices. The contribution levels \textit{Task}, \textit{Rewrite-LLM}, and \textit{Humanize} are never considered acceptable.
Across all detectors, we observe a decline in both ROC-AUC and TPR@5\%FPR as the boundary shifts toward less human contribution. When only fully human-written essays are considered acceptable, detectors achieve high ROC-AUC scores (0.87--0.94) and maintain moderate to high true positive rates.
However, once LLM-assisted revisions of human-written texts (\textit{Improve-Human} and \textit{Rewrite-Human}) are included in the negative class, TPR@5\%FPR drops across all detectors, often approaching zero.
This is most noticeable for RoBERTa, which outperforms the other detectors under the most restrictive boundary but degrades rapidly at less restrictive boundaries, suggesting high sensitivity to text characteristics introduced by LLMs.
Looking at the ROC curves in Figure~\ref{fig:boundary_roc}, the curves flatten considerably the further the boundary is shifted towards less human contribution, especially for low false positive rate regions. This indicates substantial overlap between detector score distributions of acceptable and non-acceptable texts, implying that no threshold yields meaningful recall without incurring high rates of false positives.
Overall, these results suggest that policies allowing some degree of LLM usage pose substantial challenges for current detectors, particularly when low false positive rates are crucial.

For all subsequent experiments, we assume that the policy would deem the contribution levels \textit{Human}, \textit{Improve-Human}, and \textit{Rewrite-Human} as acceptable, while all remaining levels are considered non-acceptable, reflecting typical educational policies which almost never ban LLM use entirely \cite{Wang2024}.

\subsection{Detector Performance Across Contribution Levels}
\label{sec:detector_comparison}
Assuming \textit{Rewrite-Human} as the boundary between acceptable and non-accept-\allowbreak able texts, we analyze detector performance for individual contribution levels.
Accordingly, \textit{Human}, \textit{Improve-Human}, and \textit{Rewrite-Human} are evaluated against \textit{Task} as the positive class, respectively. Conversely, \textit{Summary}, \textit{Task+Summary}, \textit{Task}, \textit{Rewrite-LLM}, and \textit{Humanize} are evaluated against \textit{Human} as the negative class (which implies that the results for Human and Task are the same).

Table~\ref{tab:detector_results_overview} reports ROC-AUC and TPR@5\%FPR scores across all detectors and contribution levels.
\begin{table*}[tb]
    \centering
    \small
    \caption{ROC-AUC and TPR@5\%FPR scores across contribution levels with \textit{Rewrite-Human} as label boundary.}
    \label{tab:detector_results_overview}
    \input{tables/contribution_level}
\end{table*}
The classification between \textit{Human} and \textit{Task} level texts is the easiest for almost all detectors, with ROC-AUC and TPR@5\%FPR reaching 0.998 for the RoBERTa model.
In contrast, performance degrades substantially at intermediate contribution levels, with \textit{Rewrite-Human} appearing as the most challenging level.
At this level, all detectors exhibit strong drops in both ROC-AUC and TPR@5\%FPR, indicating limited discriminative ability under realistic collaborative writing practices. 
While zero-shot methods exhibit a minor advantage, TPR@5\%FPR remains below 0.5 for all detectors.
Among the evaluated methods, Fast-DetectGPT retains comparatively higher performance across several intermediate contribution levels, though it struggles with LLM-assisted revisions of human-written text.
Interestingly, detector performance is less affected for the \textit{Rewrite-LLM} and \textit{Humanize} contribution levels, indicating that detectors are somewhat robust to LLM rewrites and the DIPPER humanize approach.
Overall, these results demonstrate that current detectors primarily succeed when texts are exclusively written by humans or LLMs, but fail to reliably distinguish texts produced through a human-AI collaboration.

\subsection{Detection Performance Across Generative Models}
In educational settings, student submissions may originate from a wide range of unknown generative models.
We investigate how detection performance varies when considering texts generated by a single model compared to texts generated by multiple models.
Table~\ref{tab:llm_impact} reports ROC-AUC and TPR@5\%FPR for texts generated by GPT-4o-mini, Llama-3.3, and a combination of both models.
Across all detectors, texts generated by GPT-4o-mini are consistently more difficult to detect than those generated by Llama-3.3.
On average, ROC-AUC decreases by 0.03 and TPR@5\%FPR by 0.06 for GPT-4o-mini-generated texts, indicating generator-dependent variation in detection performance.
When texts from both generative models are combined, performance further decreases for all detectors across both ROC-AUC and TPR@5\%FPR.
Because this setting reflects realistic educational scenarios with unknown and heterogeneous generative models, these findings suggest that detector performance measured under single-generator assumptions may overestimate real-world performance.

\begin{table}[tb]
    \centering
    \small
    \caption{ROC-AUC and TPR@5\%FPR across different generative models.}
    \label{tab:llm_impact}
    \input{tables/llm_impact}
\end{table}

\subsection{Detection Performance Across Educational Writing Corpora}

Writing in educational contexts differs substantially in genre, style, and task formulation.
In realistic deployment scenarios, detectors are therefore likely to encounter texts from domains that differ from those in their training data.
We assess how detection performance varies across the three GEDE corpora: AAE, BAWE, and PERSUADE.

Table~\ref{tab:dataset_comparison} reports ROC-AUC and TPR@5\%FPR scores for all detectors on each corpus.
For the supervised RoBERTa detector, we report results for all three variants, where the subscript denotes the corpus on which the model was trained.
Overall, detector performance varies noticeably across the different corpora, indicating a strong dependence on corpus-specific characteristics.
RoBERTa's performance is highly sensitive to the choice of training data.
RoBERTa$_{\text{AAE}}$ achieves the most robust results across all evaluation corpora, outperforming RoBERTa models trained on BAWE even when evaluated on those corpora.
In contrast, RoBERTa$_{\text{BAWE}}$ exhibits notably weaker performance on unseen data, despite BAWE being the largest and most diverse corpus. Although Ghostbuster is not trained on either of these text corpora, it outperforms RoBERTa$_{\text{BAWE}}$ on all of them.
These findings suggest that dataset properties are more important for cross-corpus robustness than dataset size alone, which aligns with He et al.~\cite{He2024}.

\noindent
Zero-shot detectors exhibit more stable performance across corpora, while still showing corpus-dependent differences.
Fast-DetectGPT consistently achieves the highest ROC-AUC and TPR@5\%FPR across all datasets, while DetectGPT exhibits moderate but corpus-dependent variation.
Both detectors achieve lower TPR@5\%FPR scores on the PERSUADE corpus compared to AAE and BAWE, indicating increased detection difficulty on this dataset.

\begin{table}[t]
    \centering
    \small
    \caption{ROC-AUC and TPR@5\%FPR on different text corpora.}
    \label{tab:dataset_comparison}
    \input{tables/dataset_comparison}
\end{table}

\subsection{Detection Performance under Reduced Text Length}

To analyze the impact of text length, we evaluate all detectors on truncated essays with maximum lengths of 250, 200, 150, 100, and 50 words, using the original essays of approximately 300 words as a reference. To preserve semantic and syntactic validity, truncation is applied only at sentence boundaries.
\begin{figure*}[ht]
    \centering
    \includegraphics[width=\textwidth]{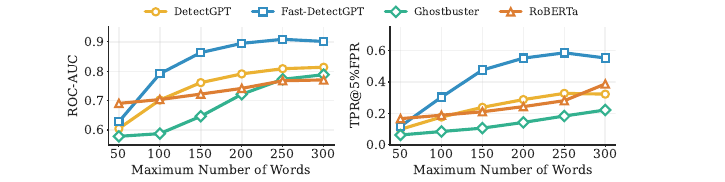}
    \caption{Detection performance across varying text lengths for ROC AUC on the left and TPR@5\%FPR on the right.}
    \label{fig:text_length}
\end{figure*}

Results are shown in Figure~\ref{fig:text_length}, reporting ROC-AUC (left) and TPR@5\%FPR (right).
Both ROC-AUC and TPR@5\%FPR decline with decreasing text length across all detectors.
Ghostbuster is most strongly affected, exhibiting considerable performance degradation below 250 words.
While DetectGPT and Fast-DetectGPT retain comparatively strong performance on longer texts, their detection performance also declines for shorter inputs.
RoBERTa exhibits the smallest relative performance drop across decreasing text lengths.
Although detection performance at very short lengths remains very limited, RoBERTa outperforms the other detectors on text below 50 words.
These findings indicate that all evaluated detectors struggle with short texts, limiting their reliability for short-answer student submissions.

%% file: tables/human_label_boundary.tex
\setlength{\tabcolsep}{5pt}

\begin{tabularx}{.95\textwidth}{l | YY | YY | YY | YY}
\toprule
 & \multicolumn{2}{l|}{DetectGPT} & \multicolumn{2}{l|}{Fast-DetectGPT} & \multicolumn{2}{l|}{Ghostbuster} & \multicolumn{2}{l}{RoBERTa} \\
Boundary & AUC & TPR & AUC & TPR & AUC & TPR & AUC & TPR \\
\midrule
Human & .931 & .615 & .930 & .735 & .865 & .574 & \textbf{.968} & \textbf{.906} \\
Improve-Human & .838 & .422 & \textbf{.891} & \textbf{.628} & .805 & .291 & .859 & .428 \\
Rewrite-Human & .814 & .323 & .\textbf{901} & .\textbf{554} & .789 & .223 & .771 & .388 \\
Summary & .682 & .087 & \textbf{.778} & .092 & .698 & \textbf{.128} & .655 & .114 \\
Task+Summary & .623 & .073 & \textbf{.724} & .065 & .672 & \textbf{.107} & .581 & .082 \\
\bottomrule
\end{tabularx}

%% file: tables/contribution_level.tex
\setlength{\tabcolsep}{2pt}
\begin{tabular}{l | cc | cc | cc | cc | cc | cc | cc | cc}
\toprule
& \multicolumn{2}{>{\columncolor{humanbg}}l|}{\textbf{Human}}
& \multicolumn{2}{>{\columncolor{humanbg}}l|}{\makecell[bl]{Improve-\\Human}}
& \multicolumn{2}{>{\columncolor{humanbg}}l|}{\makecell[bl]{Rewrite-\\Human}}
& \multicolumn{2}{>{\columncolor{llmbg}}l|}{Summary}
& \multicolumn{2}{>{\columncolor{llmbg}}l|}{\makecell[bl]{Task+\\Summary}}
& \multicolumn{2}{>{\columncolor{llmbg}}l|}{\textbf{Task}}
& \multicolumn{2}{>{\columncolor{llmbg}}l|}{\makecell[bl]{Rewrite-\\LLM}}
& \multicolumn{2}{>{\columncolor{llmbg}}l}{Humanize} \\
Detector & AUC & TPR & AUC & TPR & AUC & TPR & AUC & TPR & AUC & TPR & AUC & TPR & AUC & TPR & AUC & TPR \\
\midrule
DetectGPT & .980 & .861 & .896 & .586 & .834 & .396 & .976 & .838 & .968 & .762 & .980 & .861 & .945 & .655 & .926 & .546 \\
Fast-DetectGPT & .983 & .916 & \textbf{.926} & \textbf{.694} & \textbf{.869} & \textbf{.501} & .978 & .881 & .975 & .864 & .983 & .916 & .967 & .840 & \textbf{.996} & \textbf{.980} \\
Ghostbuster & .940 & .759 & .830 & .302 & .761 & .198 & .935 & .728 & .906 & .624 & .940 & .759 & .916 & .653 & .910 & .661 \\
RoBERTa & \textbf{.998} & \textbf{.998} & .876 & .518 & .710 & .304 & \textbf{.998} & \textbf{.998} & \textbf{.996} & \textbf{.991} & \textbf{.998} & \textbf{.998} & \textbf{.997} & \textbf{.995} & .920 & .727 \\
\bottomrule
\end{tabular}

%% file: tables/llm_impact.tex
\setlength{\tabcolsep}{3pt}
\begin{tabular}{l | c|c | c|c | c|c}
\toprule
 & \multicolumn{2}{l|}{GPT-4o-mini} & \multicolumn{2}{l|}{Llama-3.3} & \multicolumn{2}{l}{Both} \\
Detector & AUC & TPR & AUC & TPR & AUC & TPR\\
\midrule
DetectGPT & .824 & .341 & .871 & .422 & .814 & .323 \\
Fast-DetectGPT & \textbf{.931} & \textbf{.689 }& \textbf{.948} & \textbf{.753 }& \textbf{.901} & \textbf{.554} \\
Ghostbuster & .787 & .192 & .835 & .306 & .789 & .223 \\
RoBERTa & .803 & .430 & .809 & .394 & .771 & .388 \\
\bottomrule
\end{tabular}

%% file: tables/dataset_comparison.tex
\setlength{\tabcolsep}{3pt}
\begin{tabular}{l | cc | cc | cc}
\toprule
 & \multicolumn{2}{c|}{AAE} & \multicolumn{2}{c|}{BAWE} & \multicolumn{2}{c}{PERSUADE} \\
Detector & AUC & TPR & AUC & TPR & AUC & TPR \\
\midrule
DetectGPT & .840 & .388 & .806 & .329 & .772 & .196 \\
FastDetectGPT & \textbf{.880} & \textbf{.494} & \textbf{.928} & \textbf{.672} & .844 & .305 \\
Ghostbuster & .792 & .240 & .861 & .437 & .745 & .086 \\
Roberta$_\text{AAE}$ & .803 & .227 & .875 & .600 & .850 & \textbf{.357} \\
Roberta$_\text{BAWE}$ & .560 & .106 & .797 & .141 & .662 & .259 \\
Roberta$_\text{PERSUADE}$ & .702 & .098 & .798 & .242 & \textbf{.904} & .301 \\

\bottomrule
\end{tabular}

%% file: sections/06-discussion.tex
\section{Discussion}
\label{sec:discussion}

This work investigates whether current LLM-generated text detectors are suitable for use in realistic educational settings.
To this end, we introduce the concept of \emph{student contribution} and evaluate detection performance across a spectrum of human-AI collaboration practices that reflect most modern student writing.
We show that while the evaluated detectors achieve good performance under idealized conditions (exclusively human-written or exclusively LLM-generated texts), their effectiveness degrades at intermediate contribution levels, particularly for LLM-assisted rewrites of human-written text.
These findings directly challenge the common binary framing of LLM detection and indicate a misalignment between current detection approaches and institutional usage policies that permit limited LLM support \cite{Wang2024}.
Even minor LLM-assisted improvements of human-written text increase false positive rates, raising concerns about the reliability and fairness of detector-based assessments in educational practice.
We demonstrate that detector performance is further constrained by largely unavoidable factors in real-world deployment:
Detection performance varies across generative models, and performance measured under single-model assumptions overestimates performance when multiple generators are present.
Similarly, variation across educational writing corpora leads to notable performance differences, even for zero-shot detectors that are not trained on corpus-specific data.
Finally, detection performance degrades for short texts, limiting deployment in settings where student submissions consist of short answers.

While our evaluation covers a diverse set of prompts, text corpora, and generative models, some limitations should be considered when interpreting the results.
In particular, our data is limited to English-language essays, excluding other languages and educational tasks, such as programming, mathematical proofs, experimental protocols, and many more.
In addition, we used generated summaries of human-written text rather than actual human-written summaries or bullet points. Although this affects only two contribution levels, future research should validate the findings for these specific levels using real human bullet-point notes.
While this study intentionally omits intermediate contribution levels from model training to reflect current detection training practices, we strongly recommend including all levels during training to potentially improve reliability at intermediate levels.

Despite these limitations, our findings show that current detection methods are highly sensitive to even minor LLM-assisted modifications of human-written text.
These results emphasize the need to evaluate detection systems in the context of realistic institutional usage policies and to model student writing as a mix of human and LLM contributions rather than as a strict binary.

%% file: sections/09-appendix.tex
\section{Dataset Statistics}
\label{sec:dataset_statistics}
This section provides supplementary statistics of the corpora used in our experiments. 
The presented analyses aim to give a more detailed overview of the corpus characteristics underlying the GEDE dataset.

\subsection{Text Corpus Statistics}
\begin{figure}[H]
    \centering
    
    \begin{subfigure}{.49\textwidth}
        \includegraphics[width=\linewidth]{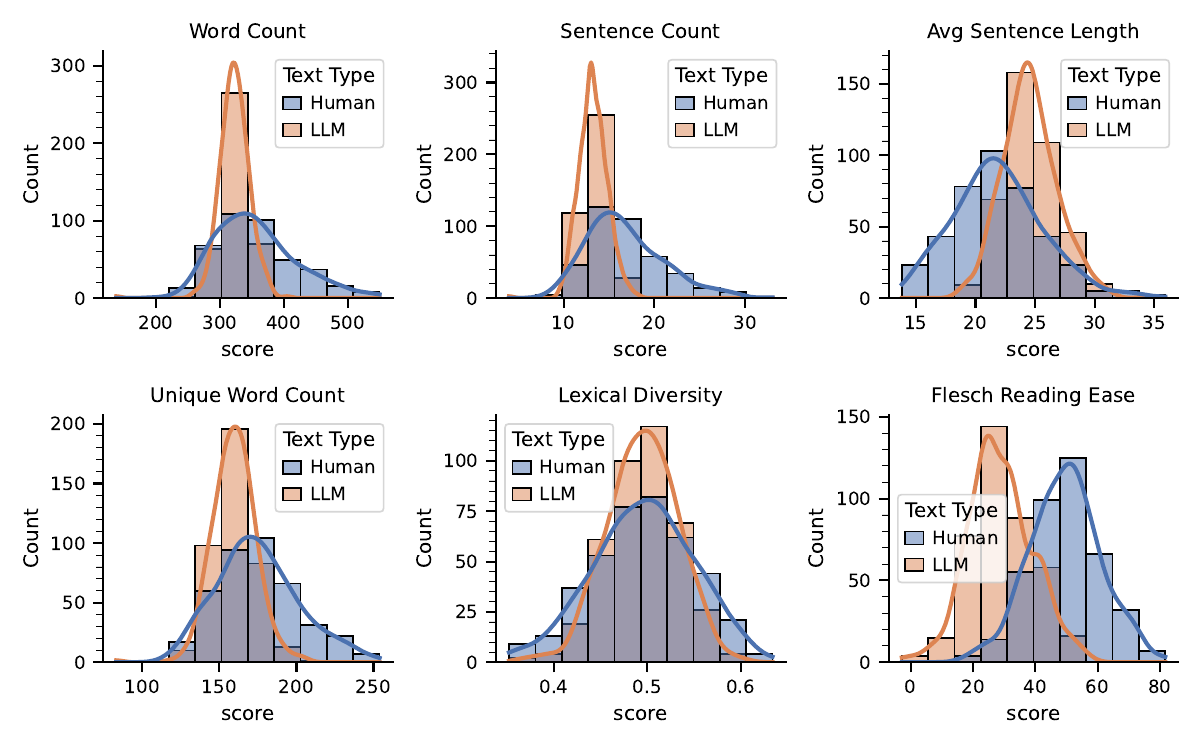}
        \caption{Llama-3.3-70b-Instruct}
    \end{subfigure}
    \hfill
    \begin{subfigure}{.49\textwidth}
        \includegraphics[width=\linewidth]{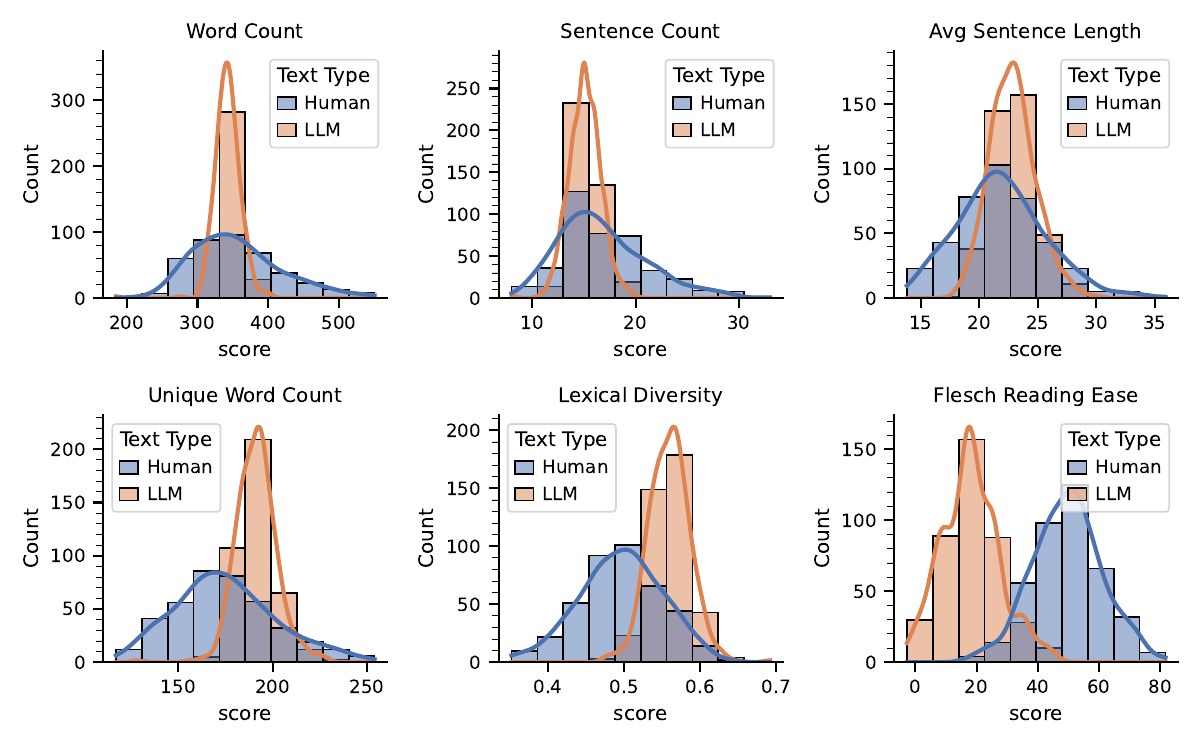}
        \caption{GPT-4o-mini}
    \end{subfigure}
    \caption{AAE \textit{Task} vs \textit{Human} statistics}
    \label{fig:aee-stats}
\end{figure}

\begin{figure}[H]
    \centering
    \begin{subfigure}{.49\textwidth}
        \includegraphics[width=\textwidth]{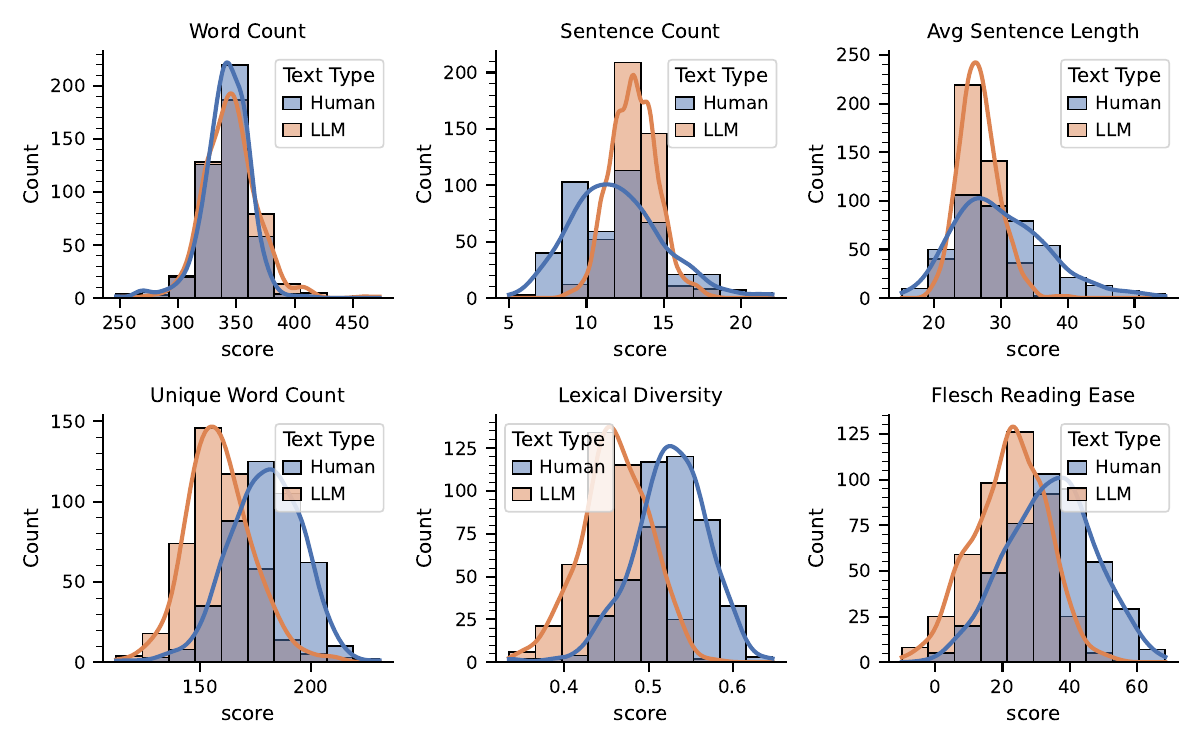}
        \caption{Llama-3.3-70b-Instruct}
    \end{subfigure}
    \hfill
    \begin{subfigure}{.49\textwidth}
        \includegraphics[width=\textwidth]{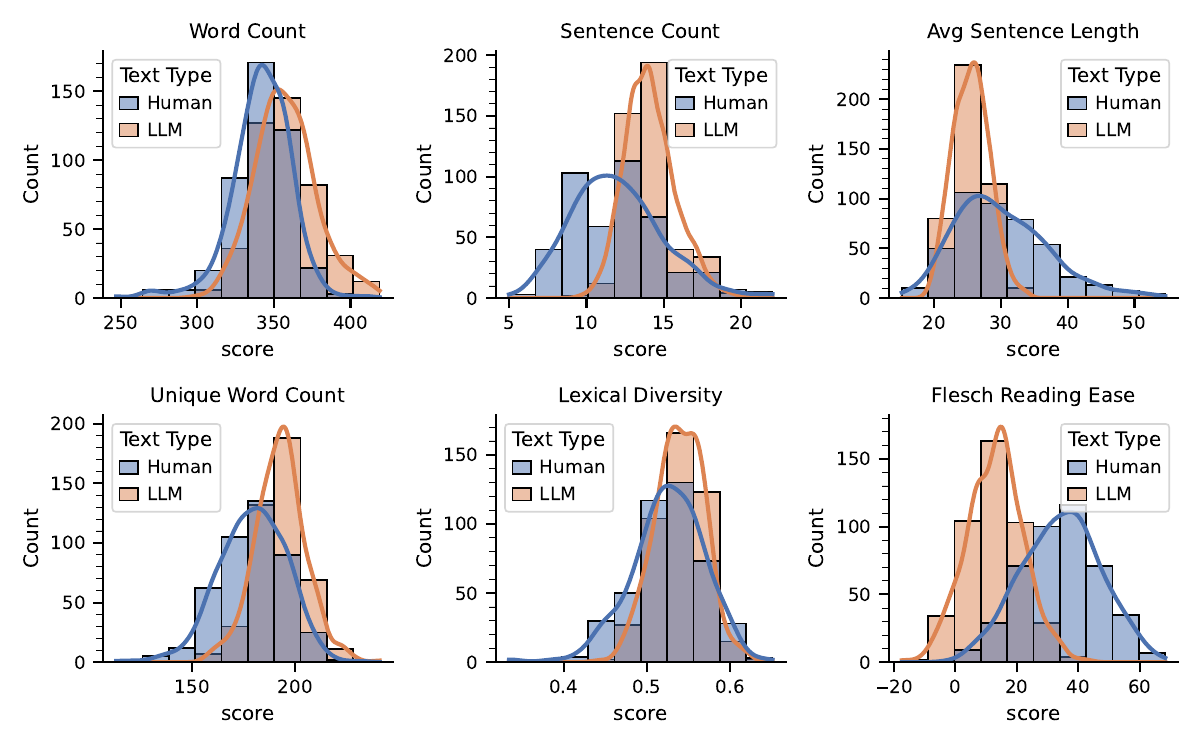}
        \caption{GPT-4o-mini}
    \end{subfigure}
    \caption{BAWE \textit{Task} vs \textit{Human} statistics}
    \label{fig:bawe-stats}
\end{figure}

\begin{figure}[H]
    \centering
    \begin{subfigure}{.49\textwidth}
        \includegraphics[width=\textwidth]{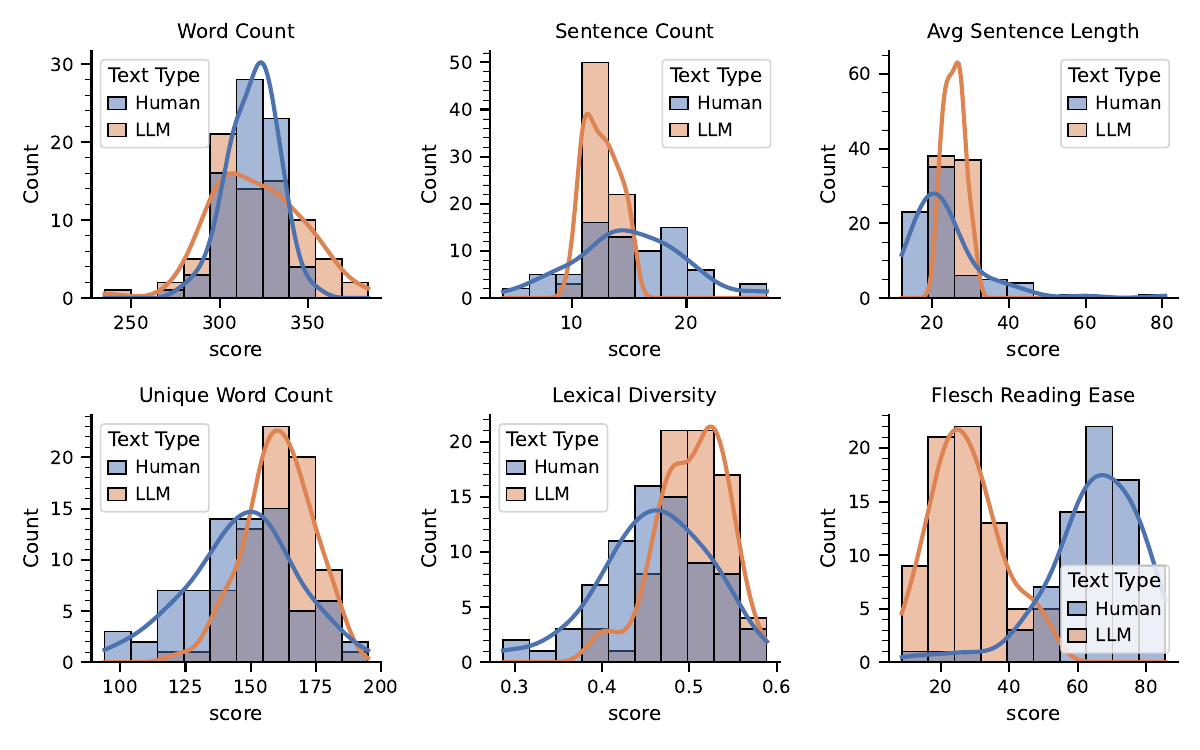}
        \caption{Llama-3.3-70b-Instruct}
    \end{subfigure}
    \hfill
    \begin{subfigure}{.49\textwidth}
        \includegraphics[width=\textwidth]{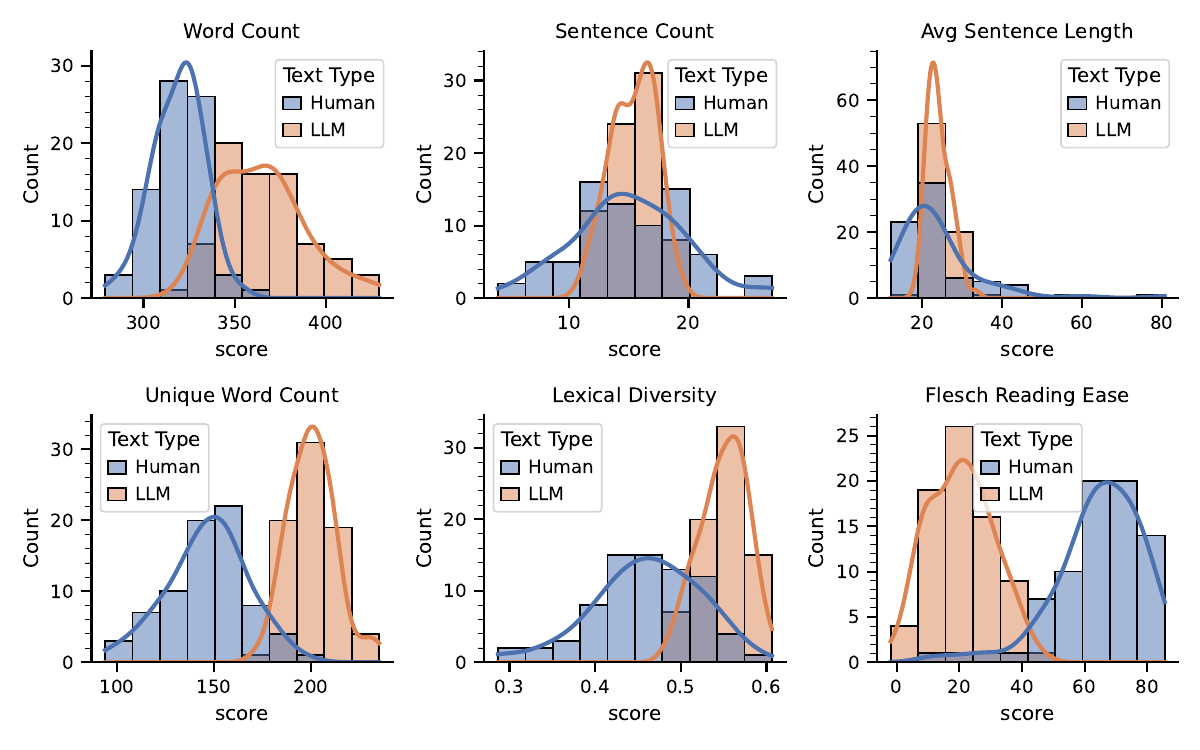}
        \caption{GPT-4o-mini}
    \end{subfigure}
    \caption{PERSUADE \textit{Task} vs \textit{Human} statistics}
    \label{fig:persuade-stats}
\end{figure}

In the following, we report detailed statistics for the AAE~\cite{stab-gurevych-2017-parsing, Stab2017} (Fig.~\ref{fig:aee-stats}), BAWE~\cite{BAWE2008} (Fig.~\ref{fig:bawe-stats}), and PERSUADE~\cite{Crossley2024} (Fig.~\ref{fig:persuade-stats}) corpora. 
For each corpus, we compare essays from the \textit{Task} contribution level with those from the \textit{Human} contribution level.

\subsection{BAWE Disciplines}
This subsection provides supplementary information on the disciplinary groups included in the BAWE corpus.
\autoref{tab:discipline-roc-comparison} summarizes the disciplinary groups, their relative frequencies, and the mean ROC-AUC scores across all detectors.

\begin{table}[H]
\caption{Discipline ROC-AUC comparison}
\label{tab:discipline-roc-comparison}
\resizebox{\columnwidth}{!}{%
\input{tables/discipline_roc_comparison}
}
\end{table}

\section{Contribution Level Prompts}
\label{sec:prompts}

This section documents the system and user prompts used to generate the LLM-based contribution levels in the GEDE dataset. 
\autoref{tab:prompts} lists all prompts employed during text generation. The \textit{Human} and \textit{Humanize} levels are included for completeness, although they are not created using system or user prompts.

\begin{table}[H]
    \small
    \caption{System and User Prompt for each contribution level.}
    \label{tab:prompts}
    \begin{tabular}{p{0.12\linewidth}p{0.4\linewidth}p{0.4\linewidth}}
        \toprule
        \textbf{Contribution \newline Level} & \textbf{System Prompt} & \textbf{User Prompt} \\
        \midrule
        Human & - & - \\
        \midrule
        Improve-Human & Your task is to improve a given text. Structure and content of the text should be retained and you should only make small improvements to the grammar and language. & 
        Rewrite this text: \{\texttt{human-text}\} \\
        \midrule

        Rewrite-Human & - & 
        Rewrite this text: \{\texttt{human-text}\} \\
        \midrule

        Summary & You are a student writing an essay on a given topic. Write around 300 words. & 
        Write an essay with the following information: \{\texttt{summary}\} \\
        \midrule

        Summary+ Task & You are a student writing an essay on a given topic. Write around 300 words. & 
        Write an essay for this task: \{\texttt{task}\} \newline Please include the following information: \{\texttt{summary}\} \\
        \midrule

        Task & You are a student writing an essay on a given topic. Write around 300 words. & 
        Write an essay for this task: \{\texttt{task}\} \\
        \midrule

        Rewrite-LLM & - & 
        Rewrite this text: \{\texttt{llm-text}\} \\
        \midrule

        Humanize & - & - \\
        \bottomrule
    \end{tabular}
\end{table}

\section{Changes to Human-Written Text}
\label{sec:human-text-changes}

This section provides qualitative examples illustrating how large language models modify human-written texts at different contribution levels.

\autoref{fig:text-changes-for-improve-human-gpt} shows an excerpt from a randomly selected human-written text from the BAWE corpus that was improved by GPT-4o-mini for the \textit{Improve-Human} contribution level. Text passages highlighted from orange to blue indicate modifications of existing content, while red denotes removed words and green newly added ones. Although the model introduces several changes, the overall meaning of the text is preserved. In comparison, the improvements produced by LLaMA-3.3 in \autoref{fig:text-changes-for-improve-human-llama} involve fewer edits to the original text. Nevertheless, both models exhibit similar editing patterns, such as replacing \textit{changes} with \textit{shift} or simplifying expressions like \textit{in order to} to \textit{to}.

\begin{figure}[H]
    \centering
    \includegraphics[width=0.5\linewidth]{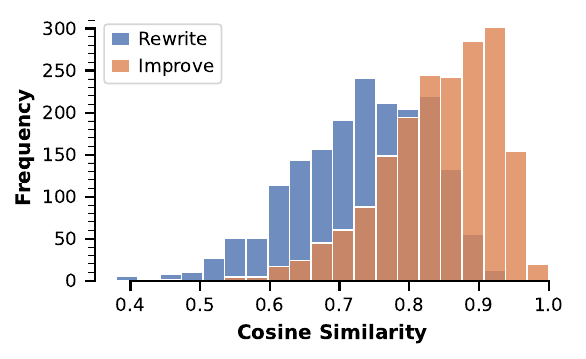}
    \caption{Cosine Similarity between Improve-Human and Rewrite-Human texts.}
    \label{fig:cos-sim}
\end{figure}

In addition, we compute the cosine similarity between the original human-written texts and their improved or rewritten counterparts from the \textit{Improve-Human} and \textit{Rewrite-Human} contribution levels. As shown in \autoref{fig:cos-sim}, texts from the \textit{Improve-Human} category are more similar to the original texts than those from the \textit{Rewrite-Human} category.

\begin{figure}[H]
    \centering
    \begin{subfigure}{\textwidth}
        \begin{framed}
            \small
            The present study aims to discuss the implications for extension practice \textcolor{replace_old}{\textbf{of changes}} $\rightarrow$ \textcolor{replace_new}{\textbf{resulting from a shift}} in thinking \textcolor{delete}{\textbf{about extension}} from 'transfer of technology' to 'communication to support \textcolor{replace_old}{\textbf{innovation'. In order to discuss}} $\rightarrow$ \textcolor{replace_new}{\textbf{innovation.' To explore}} this topic, \textcolor{replace_old}{\textbf{this}} $\rightarrow$ \textcolor{replace_new}{\textbf{the}} study will \textcolor{replace_old}{\textbf{present}} $\rightarrow$ \textcolor{replace_new}{\textbf{provide}} an overview of what innovation \textcolor{replace_old}{\textbf{is, what}} $\rightarrow$ \textcolor{replace_new}{\textbf{entails, the focus of technology transfer, and the reasons for transitioning from a technology}} transfer \textcolor{replace_old}{\textbf{of technology focus, and the reasons why there is a need to shift from transfer of technology}} $\rightarrow$ \textcolor{replace_new}{\textbf{model}} to an innovative approach \textcolor{replace_old}{\textbf{in the extension practices context. In}} $\rightarrow$ \textcolor{replace_new}{\textbf{within}} the context of \textcolor{insert}{\textbf{extension practices. For}} this study, rural extension is defined as assistance \textcolor{insert}{\textbf{provided}} to individuals living in rural areas \textcolor{replace_old}{\textbf{(farmers}} $\rightarrow$ \textcolor{replace_new}{\textbf{(whether farmers}} or not) \textcolor{replace_old}{\textbf{in helping}} $\rightarrow$ \textcolor{replace_new}{\textbf{to help}} them \textcolor{delete}{\textbf{to}} identify and \textcolor{replace_old}{\textbf{analyse}} $\rightarrow$ \textcolor{replace_new}{\textbf{analyze}} their \textcolor{replace_old}{\textbf{problems, and}} $\rightarrow$ \textcolor{replace_new}{\textbf{problems while also}} being aware of opportunities for improvement (see Adams, 1982). \textcolor{replace_old}{\textbf{Extension was usually}} $\rightarrow$ \textcolor{replace_new}{\textbf{Traditionally, extension has been}} associated with increasing food production and \textcolor{replace_old}{\textbf{encouraging}} $\rightarrow$ \textcolor{replace_new}{\textbf{promoting}} economic development, \textcolor{replace_old}{\textbf{having the function of promoting}} $\rightarrow$ \textcolor{replace_new}{\textbf{functioning as a means of disseminating}} knowledge and transferring technology between farmers and researchers (or \textcolor{insert}{\textbf{among}} farmers \textcolor{replace_old}{\textbf{to farmers)}} $\rightarrow$ \textcolor{replace_new}{\textbf{themselves)}} (Leeuwis, 2004:17). ...
        \end{framed}
        \caption{Visualization of the changes made by GPT-4o-mini.}
        \label{fig:text-changes-for-improve-human-gpt}
    \end{subfigure}
    
    \begin{subfigure}{\textwidth}
        \begin{framed}
            \small
            The present study aims to \textcolor{replace_old}{\textbf{discuss}} $\rightarrow$ \textcolor{replace_new}{\textbf{explore}} the implications for extension practice of \textcolor{replace_old}{\textbf{changes}} $\rightarrow$ \textcolor{replace_new}{\textbf{the shift}} in thinking about extension from 'transfer of technology' to 'communication to support innovation'. \textcolor{replace_old}{\textbf{In order to}} $\rightarrow$ \textcolor{replace_new}{\textbf{To}} discuss this topic, this study will \textcolor{replace_old}{\textbf{present}} $\rightarrow$ \textcolor{replace_new}{\textbf{provide}} an overview of what innovation \textcolor{replace_old}{\textbf{is, what}} $\rightarrow$ \textcolor{replace_new}{\textbf{entails, the focus of}} transfer of \textcolor{replace_old}{\textbf{technology focus,}} $\rightarrow$ \textcolor{replace_new}{\textbf{technology,}} and the reasons why \textcolor{replace_old}{\textbf{there is a need to}} $\rightarrow$ \textcolor{replace_new}{\textbf{a}} shift from transfer of technology to an innovative approach \textcolor{replace_old}{\textbf{in the extension practices context.}} $\rightarrow$ \textcolor{replace_new}{\textbf{is necessary in the context of extension practices.}} In \textcolor{delete}{\textbf{the context of}} this study, rural extension is defined as assistance to individuals living in rural areas \textcolor{replace_old}{\textbf{(farmers}} $\rightarrow$ \textcolor{replace_new}{\textbf{(whether farmers}} or not) in \textcolor{replace_old}{\textbf{helping them to identify and analyse}} $\rightarrow$ \textcolor{replace_new}{\textbf{identifying and analyzing}} their problems, and \textcolor{replace_old}{\textbf{being}} $\rightarrow$ \textcolor{replace_new}{\textbf{becoming}} aware of opportunities for improvement (see Adams, 1982). \textcolor{replace_old}{\textbf{Extension}} $\rightarrow$ \textcolor{replace_new}{\textbf{Traditionally, extension}} was \textcolor{delete}{\textbf{usually}} associated with increasing food production and \textcolor{replace_old}{\textbf{encouraging}} $\rightarrow$ \textcolor{replace_new}{\textbf{promoting}} economic development, \textcolor{replace_old}{\textbf{having the}} $\rightarrow$ \textcolor{replace_new}{\textbf{with the primary}} function of \textcolor{replace_old}{\textbf{promoting}} $\rightarrow$ \textcolor{replace_new}{\textbf{disseminating}} knowledge and transferring technology between farmers and researchers (or \textcolor{insert}{\textbf{among}} farmers \textcolor{replace_old}{\textbf{to farmers)}} $\rightarrow$ \textcolor{replace_new}{\textbf{themselves)}} (Leeuwis, 2004:17). ...
        \end{framed}
        \caption{Visualization of the changes made by Llama-3.3-70b-Instruct.}
        \label{fig:text-changes-for-improve-human-llama}
    \end{subfigure}
    \caption{Visualization of the changes made by both LLMs to a human-written text from the BAWE subset, belonging to the \textit{Improve-Human} contribution level.}
    \label{fig:text-changes-for-improve-human}
\end{figure}

\section{Large Language Model Configurations}
Table~\ref{tab:model_settings} summarizes the configurations of the language models used to generate the essays in the GEDE dataset. 
It provides an overview of the model versions and the most relevant inference parameters to ensure transparency and reproducibility.
\begin{table}[H]
\centering
\caption{Overview of LLM model configurations}
\label{tab:model_settings}
\begin{tabular}{lcc}
\toprule
\textbf{Parameter} & \textbf{GPT-4o-mini} & \textbf{Llama-3.3-70B} \\
\midrule
Model version & gpt-4o-mini-2024-07-18 & Llama-3.3-70b-Instruct \\
Provider / Source & OpenAI API & Hugging Face \\
Quantization & -- & 4-bit \\
Temperature & 1.0 & 1.0 \\
Max new tokens & 512 & 512 \\
\bottomrule
\end{tabular}
\end{table}

\section{RoBERTa Fine-Tuning Details}
\label{sec:roberta-training}
We fine-tune the RoBERTa model using the HuggingFace \textit{roberta-base} model\footnote{https://huggingface.co/FacebookAI/roberta-base} on each of the three text corpora. \autoref{tab:roberta-training} shows the hyperparameters we used for fine-tuning RoBERTa on the different subsets. We chose the best epoch based on the evaluation cross-entropy loss. All experiments were conducted with a fixed random seed of 42 to ensure reproducibility.

\begin{table}[H]
    \centering
    \caption{Hyperparameters of RoBERTa fine-tuning on the different text corpora subsets.}
    \label{tab:roberta-training}
    \small
    \begin{tabular}{lcccc}
        \toprule
        Subset & Batch Size & Test Size & Epochs & Best Epoch \\
        \midrule
        AAE & 32 & 0.2 & 5 & 2\\ 
        BAWE & 32 & 0.2 & 5 & 4\\ 
        PERSUADE & 32 & 0.2 & 8 & 4\\ 
        \bottomrule
    \end{tabular}
\end{table}

\section{Supplementary ROC Curves}
\label{sec:roc_curves}
This section provides additional ROC curves corresponding to the experiments described in Sections~4.2--4.5 of the main paper.

\subsection{Contribution Level}
\begin{figure}[H]
    \centering
    \includegraphics[width=\linewidth]{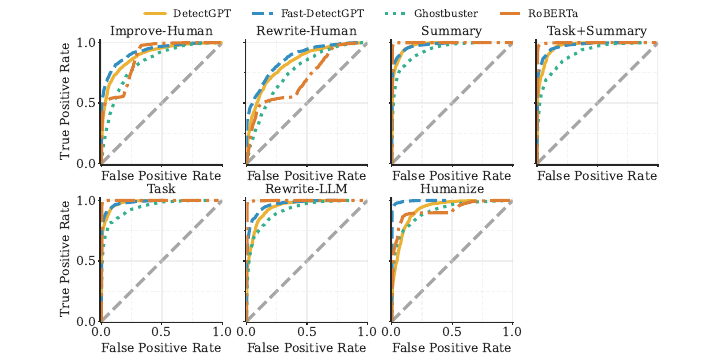}
    \caption{ROC curves for the contribution-level experiment.}
    \label{fig:roc-contribution-level}
\end{figure}

The ROC curves shown in \autoref{fig:roc-contribution-level} correspond to the contribution-level experiment described in Section~4.2.

\subsection{Generative Model}
\begin{figure}[H]
    \centering
    \includegraphics[width=\linewidth]{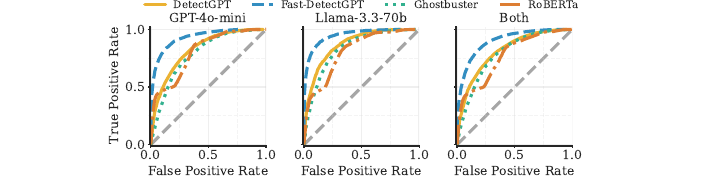}
    \caption{ROC curves for the generative-model experiment.}
    \label{fig:roc-generative-model}
\end{figure}

The ROC curves shown in \autoref{fig:roc-generative-model} correspond to the generative-model experiment described in Section~4.3.

\subsection{Out-of-Distribution Data}
\begin{figure}[H]
    \centering
    \includegraphics[width=\linewidth]{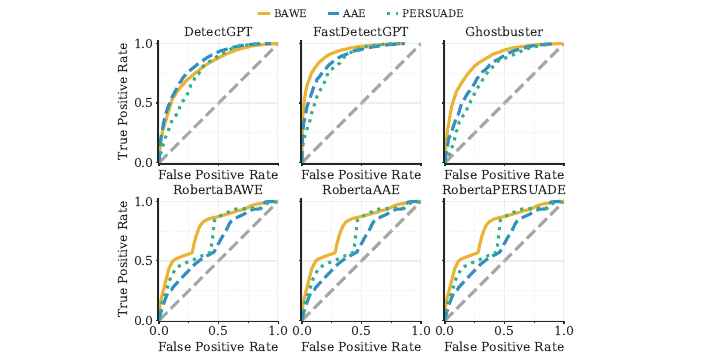}
    \caption{ROC curves for the out-of-distribution data experiment.}
    \label{fig:roc-ood}
\end{figure}

The ROC curves shown in \autoref{fig:roc-ood} correspond to the out-of-distribution data experiment described in Section~4.4.

\subsection{Text Lenght}
\begin{figure}[H]
    \centering
    \includegraphics[width=\linewidth]{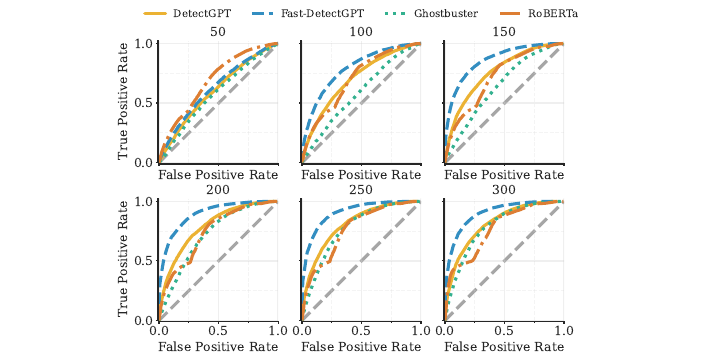}
    \caption{ROC curves for the text length experiment.}
    \label{fig:roc-length}
\end{figure}

The ROC curves shown in \autoref{fig:roc-length} correspond to the text length experiment described in Section~4.5.

\newpage

%% file: tables/discipline_roc_comparison.tex
\begin{tabular}{lcccccc}
\toprule
discipline\_group & Group Mean ± Std & group\_cout & discipline & ROC-AUC (Mean ± Std) & dis\_cout \\
\midrule
\multirow{9}{*}{AH} & \multirow{9}{*}{0.812 ± 0.11} & \multirow{9}{*}{197} & History & 0.841 ± 0.09 & 25 \\
 &  &  & Comparative American Studies & 0.841 ± 0.09 & 25 \\
 &  &  & Other & 0.841 ± 0.08 & 22 \\
 &  &  & Archaeology & 0.829 ± 0.13 & 25 \\
 &  &  & English & 0.823 ± 0.10 & 25 \\
 &  &  & Classics & 0.804 ± 0.11 & 25 \\
 &  &  & Linguistics & 0.793 ± 0.15 & 24 \\
 &  &  & OTHER & 0.786 ± 0.17 & 1 \\
 &  &  & Philosophy & 0.753 ± 0.15 & 25 \\
 \midrule
\multirow{6}{*}{LS} & \multirow{6}{*}{0.823 ± 0.14} & \multirow{6}{*}{57} & Medicine & 0.868 ± 0.10 & 4 \\
 &  &  & Biological Sciences & 0.864 ± 0.07 & 4 \\
 &  &  & Agriculture & 0.823 ± 0.18 & 11 \\
 &  &  & Psychology & 0.805 ± 0.17 & 25 \\
 &  &  & Health & 0.802 ± 0.15 & 10 \\
 &  &  & Food Sciences & 0.775 ± 0.15 & 3 \\
 \midrule
\multirow{5}{*}{PS} & \multirow{5}{*}{0.830 ± 0.13} & \multirow{5}{*}{16} & Chemistry & 0.910 ± 0.06 & 1 \\
 &  &  & Computer Science & 0.880 ± 0.09 & 3 \\
 &  &  & Architecture & 0.816 ± 0.14 & 3 \\
 &  &  & Planning & 0.812 ± 0.13 & 6 \\
 &  &  & Physics & 0.731 ± 0.16 & 3 \\
 \midrule
\multirow{9}{*}{SS} & \multirow{9}{*}{0.828 ± 0.12} & \multirow{9}{*}{169} & Publishing & 0.859 ± 0.08 & 2 \\
 &  &  & HLTM & 0.858 ± 0.11 & 17 \\
 &  &  & Business & 0.829 ± 0.16 & 25 \\
 &  &  & Anthropology & 0.828 ± 0.15 & 22 \\
 &  &  & Sociology & 0.826 ± 0.14 & 25 \\
 &  &  & Law & 0.819 ± 0.10 & 25 \\
 &  &  & Other & 0.815 ± 0.15 & 3 \\
 &  &  & Economics & 0.814 ± 0.13 & 25 \\
 &  &  & Politics & 0.804 ± 0.14 & 25 \\
\bottomrule
\end{tabular}